\documentclass[journal]{vgtc}                

\ifpdf
  \pdfoutput=1\relax                   
  \usepackage{graphicx}                
  \DeclareGraphicsExtensions{.pdf,.png,.jpg,.jpeg} 
\else
  \usepackage{graphicx}                
  \DeclareGraphicsExtensions{.eps}     
\fi%

\graphicspath{{figures/}{pictures/}{images/}{./}} 

\usepackage{microtype}                 
\PassOptionsToPackage{warn}{textcomp}  
\usepackage{textcomp}                  
\usepackage{mathptmx}                  
\usepackage{times}                     
\usepackage{cite}                      
\usepackage{tabu}                      
\usepackage{booktabs}                  
\usepackage{graphicx}
\usepackage{tikz}
\usepackage{pgfplots}
\usepackage[normalem]{ulem}
\usetikzlibrary{matrix}
\usepgfplotslibrary{groupplots}
\pgfplotsset{compat=newest}
\usepackage{pgfplotstable}
\usepackage{amsmath}
\usepackage{float}
\usepackage{amsfonts}
\floatstyle{plaintop}
\restylefloat{table}
\usepackage{caption}
\onlineid{0}

\vgtccategory{Research}
\vgtcpapertype{algorithm/technique}

\title{GPGPU Linear Complexity t-SNE Optimization}

\definecolor{ggreen}{rgb}{0.60, 0.5, 0.0}

\newcommand{\revDeleted}[1]{}
\newcommand{\revNew}[1]{#1}
\newcommand{\revNotes}[1]{}

\author{Nicola Pezzotti*, Julian Thijssen*, Alexander Mordvintsev, Thomas H\"ollt,\\Baldur van Lew, Boudewijn P.F. Lelieveldt, Elmar Eisemann and Anna Vilanova}
\authorfooter{
\item
 Nicola Pezzotti and Alexander Mordvintsev are with Google AI, Z\"urich, Switzerland. E-mail: nicola.pezzotti@gmail.com.
\item
 Nicola Pezzotti, Julian Thijssen, Thomas H\"ollt, Boudewijn P.F. Lelieveldt, Elmar Eisemann and Anna Vilanova are with the Delft University of Technology, Delft, The Netherlands.
\item
 Thomas H\"ollt, Baldur van Lew and Boudewijn P.F. Lelieveldt are with the Leiden University Medical Center, Leiden, The Netherlands.
}

\shortauthortitle{Biv \MakeLowercase{\textit{et al.}}: Global Illumination for Fun and Profit}

\abstract{In recent years the t-distributed Stochastic Neighbor Embedding (t-SNE) algorithm has become one of the most used and insightful techniques for exploratory data analysis of high-dimensional data.
It reveals clusters of high-dimensional data points at different scales while only requiring minimal tuning of its parameters.
\revDeleted{Despite this}\revNew{However}, the computational complexity of the algorithm limits its application to relatively small datasets. 
To address this problem, several evolutions of t-SNE have been developed in recent years, mainly focusing on the scalability of the similarity computations between data points. 
However, these contributions are insufficient to achieve interactive rates when visualizing the evolution of the t-SNE embedding for large datasets.
In this work, we present a novel approach to the minimization of the t-SNE objective function that heavily relies on \revDeleted{modern} graphics hardware and has linear computational complexity.
Our technique decreases the computational cost of running t-SNE on datasets by orders of magnitude and retains or improves on the accuracy of past approximated techniques.
We propose to approximate the repulsive forces between data points by splatting kernel textures for each data point.
This approximation allows us to reformulate the t-SNE minimization problem as a series of tensor operations that can be efficiently executed on the graphics card.
An efficient implementation of our technique is integrated and available for use in the widely used Google TensorFlow.js, and an open-source C++ library.} 


\CCScatlist{ 
 \CCScat{K.6.1}{Management of Computing and Information Systems}%
{Project and People Management}{Life Cycle};
 \CCScat{K.7.m}{The Computing Profession}{Miscellaneous}{Ethics}
}

\teaser{
  \centering
  \includegraphics[width=0.75\linewidth]{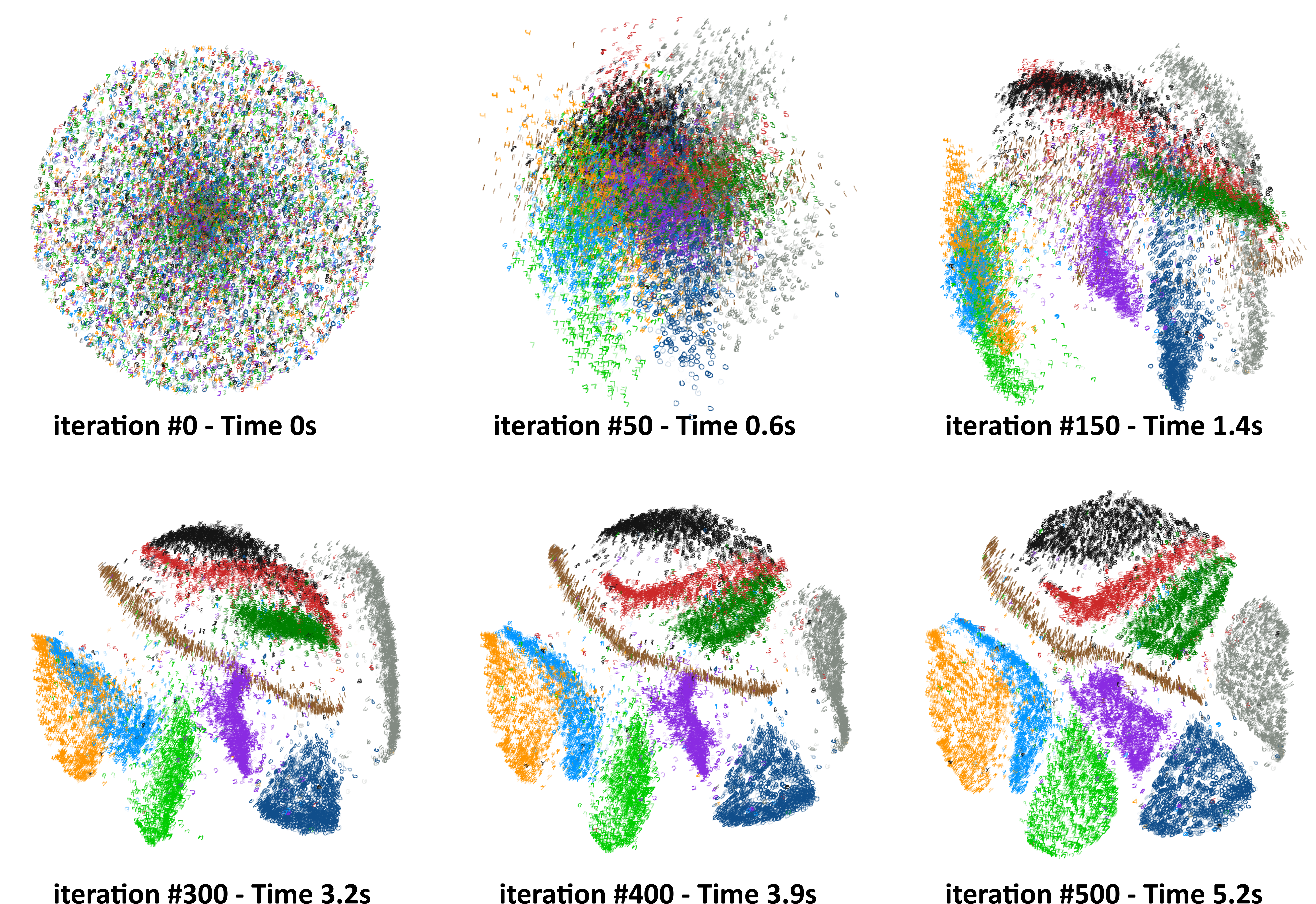}
  \caption{Evolution of the t-SNE embedding for the MNIST dataset. The optimization is performed in only a few seconds while running in a web browser and providing progressive updates. Previous implementations require tens of minutes to run in multithreaded C++ programs. CUDA implementations exists, but require NVIDIA GPUs and do not run in the browser. The example can be run at the following link \url{https://nicola17.github.io/tfjs-tsne-demo/}}
	\label{fig:teaser_new}
}

\newcommand{\manuscriptnotetxt}{
Manuscript received xx xxx. 201x; accepted xx xxx. 201x. Date of Publication xx xxx. 201x; date of current version xx xxx. 201x. For information on obtaining reprints of this article, please send  e-mail to: reprints@ieee.org. Digital Object Identifier: xx.xxxx/TVCG.201x.xxxxxxx
}

\vgtcinsertpkg

\begin{document}


\firstsection{Introduction}

\maketitle

Understanding how data points are arranged in a high-dimensional space plays a crucial role in exploratory data analysis~\cite{Tukey:1962:EDA}. 
In recent years, non-linear dimensionality reduction techniques became powerful tools for mining knowledge from data, such as for the discovery of clusters.
In the field of data visualization, these techniques are used for reducing the dimensionality to two or three dimensions in order to make visualization possible. Specifically, the algorithms preserve certain characteristics of the data, such as the local neighborhoods. \revDeleted{This is successful}\revNew{This is effective} due to the fact that most of the real-world data satisfy the ``manifold hypothesis'', i.e., they lie on low-dimensional manifolds embedded in high-dimensional space.

The t-distributed Stochastic Neighbor Embedding (t-SNE) algorithm~\cite{VanDerMaaten:2008:tSNE} has become one of the state-of-the-art non-linear dimensionality reduction methods for visual analysis of high-dimensional data. It has been successfully applied to different domains, such as life sciences~\cite{Amir:2013:viSNE,Becher:2014:MurineMyeloidCell,Lijem:2018:AtSNEPathways}, the comprehension of machine-learning models and to human-driven supervision~\cite{Mnih:2015:AtariDL,Pezzotti:2018:DeepEyes,Kahng2018ActiVis}.
The t-SNE algorithm can be separated in two computation modules; first it computes the similarities of the high-dimensional points as a joint probability distribution and, second, it minimizes the Kullback--Leibler (KL) divergence~\cite{Kullback:1997:InformationTheory}, which measures the similarity between the data distribution in the high-dimensional space and the low-dimensional \revDeleted{embedding }space.

The gradient of the KL divergence can be interpreted as a summation of attractive and repulsive forces between points, which makes the minimization process very similar to an N-body simulation\cite{NBodySimulation}.
The memory and computational complexity of the algorithm is $O\left(N^2\right)$, where $N$ is the number of data points. Interactive computation times are essential in an interactive visual exploration solution, and in consequence much research effort has been spent on improving its computational and memory complexity.

While many works focused on improvement of the similarity computation~\cite{VanDerMaaten:2014:BH-SNE,Pezzotti:2015:AtSNE,Tang:2016:LargeVis,Pezzotti:2016:HSNE,McInnes:2018:UMAP}, only limited effort has been spent on improving the minimization algorithm employed for the creation of the embedding~\cite{VanDerMaaten:2014:BH-SNE,McInnes:2018:UMAP,Kim:2016:PixelSNE}.
Barnes-Hut-SNE (BH-SNE) was proposed by van der Maaten~\cite{VanDerMaaten:2014:BH-SNE}. It makes use of the Barnes-Hut algorithm for N-body simulations\cite{Barnes:1986:BarnesHutAlgo} to approximate the repulsive forces between the data points. Repulsive forces change during minimization, since they depend on the data points position in the low-dimensional embedding space. 
Despite the improvements the computational costs remain high for large amounts of data points.

In this work, we focus on the minimization of the objective function, i.e., the KL-divergence, for the creation of the embedding.
We observe that the heavy tail of the \revDeleted{t-Student distribution}\revNew{Student's t-distribution} used by t-SNE makes the application of an $N$-body simulation not particularly effective.
We propose a paradigm shift from point-to-point computation to a field-based computation of the embedding by reformulating the gradient of the objective function as a function of scalar and vector fields combined with tensor operations.

Our technique has linear computational and memory complexity, $O\left(N\right)$, and is suitable for implementation in a GPGPU fashion, providing considerably better computation times compared to the current state of the art.
It also allows us to implement a version for the browser and desktop that minimizes the objective function for standard datasets in a matter of seconds, potentially enabling the development of more advanced web-based analytics solutions.

The contribution of our work is twofold:
\begin{itemize}
\item A linear complexity minimization of the t-SNE objective function.
Specifically, we
\begin{itemize}
\item approximate the repulsive forces between data points with a GPGPU approach relying on texture splatting
\item adopt a tensor-based computation of the objective function's gradient.
\end{itemize}
\item An efficient implementation of our approach is released as part of Google's TensorFlow.js library and as part of the C++ HDI library. Our implementation is not only several orders of magnitude faster than the Barnes-Hut-SNE, but we demonstrate that it minimizes the objective function more effectively in addition to  having better high-dimensional neighbor preservation.
\end{itemize}

The rest of the paper is structured as follows. 
In the next section, we provide a theoretical primer on the t-SNE algorithm that is needed to understand the related work (Section~\ref{sec:rw}) and our contributions (Section~\ref{sec:main}).
In Section~\ref{sec:impl}, we provide details regarding our implementations.
Finally, in Section~\ref{sec:res}, we compare our technique to BH-SNE, t-SNE-CUDA and the original t-SNE. We show the performance and accuracy improvements over these techniques using publicly available high-dimensional datasets.

\pagebreak
\section{\texorpdfstring{\MakeLowercase{t}}{}-SNE}
\label{sec:tSNE}
In this section, we provide an introduction to the t-SNE~\cite{VanDerMaaten:2008:tSNE} algorithm, which is essential to understand the related work and our contribution.
The t-SNE algorithm interprets the overall distances between data points in the high-dimensional space as a symmetric joint probability distribution $P$ that encodes their similarities.
Likewise a joint probability distribution $Q$ is computed that describes the similarity in the low-dimensional space.
The goal is to achieve a representation, referred to as an \emph{embedding}, in which $Q$ faithfully represents $P$. \revDeleted{The embedding preserves the local neighborhood of the high-dimensional data points, having the main advantage, in our context, to lay in 2- or 3-dimensional space that can be easily visualized.} \revNew{This means that the embedding preserves the local neighborhoods of the high-dimensional data points. At the same time, the low-dimensional embedding only has two or three dimensions, which can easily be visualized.}

This \revDeleted{goal}\revNew{objective} is achieved by optimizing the positions of the points in the low-dimensional embedding to minimize the cost function $C$ given by the Kullback--Leibler, $KL$, divergence between the joint-probability distributions $P$ and $Q$.
Intuitively, points in the embeddings are moved in an iterative fashion, such that the embedding similarities encoded by $Q$ \revDeleted{match}\revNew{become more closely matched} to the similarities in the high-dimensional space encoded by $P$.

In more detail, given two data points $\mathbf{x}_i$ and $\mathbf{x}_j$ in a high-dimensional dataset $X=\left\{\mathbf{x}_1 ... \mathbf{x}_N\right\}$, the probability $p_{ij}$ models the similarity of these points in high-dimensional space. $q_{ij}$ models the similarity in the low-dimensional embedding of the corresponding points $\mathbf{y}_i$ and $\mathbf{y}_j$. The cost function $C$ is formulated as follows:

\begin{equation}\label{eq:KL}
 C(P,Q) = KL(P||Q) = \sum_{i=1}^N\sum_{j=1, j\ne i}^N p_{ij} \ln\left(\frac{p_{ij}}{q_{ij}}\right),
\end{equation}

where $KL$ measure the mismatch between $Q$ and $P$.
Similarities between two points $\mathbf{x}_i$ and $\mathbf{x}_j$ in the high-dimensional space are represented by $p_{ij}$. More specifically, for each point $\mathbf{x}_i$, a Gaussian kernel is centered on the point and used to compute the probability that the other point is a neighbor. The variance $\sigma_i$ of the kernel is defined according to the local density in the high-dimensional space, and $p_{ij}$ is computed as follows:

\begin{equation} \label{eq:tsne_HD}
p_{ij} = \frac{p_{i|j} + p_{j|i}}{2N},
\end{equation} 

\begin{equation} \label{eq:P_j|i}
\text{where} \quad  p_{j|i}=\frac{\exp(-(||\mathbf{x}_i-\mathbf{x}_j||^2)/(2 \sigma_i^2))}{\sum_{k\ne i}^N \exp(-(||\mathbf{x}_i-\mathbf{x}_k||^2)/(2 \sigma_i^2))}
\end{equation}

$p_{j|i}$ can be seen as a relative measure of similarity for the point $\mathbf{x}_i$ and all the points $\mathbf{x}_j$ in its local neighborhood.
The effective number of neighbors considered for each data point is derived by the perplexity value $\mu$, which is a user-defined parameter.
Consequently, the value of $\sigma_i$ is chosen such that for a fixed perplexity $\mu$ and for each $i$ it satisfies:

\begin{equation} \label{eq:perplexity}
\mu = 2^{-\sum_{j}^N p_{j|i} \log_2 p_{j|i}}
\end{equation}

A \emph{Student's t-Distribution} with one degree of freedom is used to compute the joint probability distribution in the low-dimensional embedding $Q$, where the positions of the data points should be optimized. 
$Q$ plays a similar role for the points in the low-dimensional space, as $P$ does for the high-dimensional space.
It encodes the similarities given the neighborhood information. In the embedding space the dispersion of the distribution (i.e., the Student's t-Distribution) is constant. Given two low-dimensional points $\mathbf{y}_i$ and $\mathbf{y}_j$, the probability $q_{ij}$ is given by:

\begin{equation} \label{eq:Q}
q_{ij} = \left((1+||\mathbf{y}_i - \mathbf{y}_j||^2) Z \right)^{-1}
\end{equation}

\begin{equation} \label{eq:Z_normalization_factor}
\text{with} \quad Z = \sum_{k=1}^N\sum_{l \ne k}^N(1+||\mathbf{y}_k-\mathbf{y}_l||^2)^{-1}
\end{equation}

The goal of a t-SNE optimization is to move randomly initialized points $\mathbf{y_i}$ in the embedding, such that the distribution $Q$ is as close as possible to the distribution $P$.
Intuitively, when $Q$ matches $P$, the neighborhoods in the low-dimensional space match the high-dimensional counterparts.
This result is obtained by minimizing a cost function $C$ which is defined as the Kullback--Leibler divergence between $P$ and $Q$.
The gradient of $C$ has an analytical solution and indicates the change in position of the points $\mathbf{y_i}$. It is given by:

\begin{align} \label{eq:C_gradient_n_body}
\frac{\delta C}{\delta \mathbf{y}_i} &= 4 ( F^{\text{attr}}_i-F^{\text{rep}}_i)\\
\label{eq:C_gradient_n_body_detail}
&=4 ( Z\sum_{j \ne i}^N p_{ij}q_{ij}(\mathbf{y}_i-\mathbf{y}_j) - \sum_{j \ne i}^N q_{ij}^2Z(\mathbf{y}_i-\mathbf{y}_j))
\end{align}

The optimization is based on gradient descent. For each iteration, the gradient is used to update the position of the data points in the embedding.
The gradient descent can be seen as an \emph{$N$-body simulation}~\cite{NBodySimulation}, where each data point exerts an attractive and a repulsive force ($F^{\text{attr}}_i$ and $F^{\text{rep}}_i$) on all other points.

\section{Related Work}
\label{sec:rw}
We now present the work that has been done to improve the computation of t-SNE embeddings in terms of quality and scalability.
Van der Maaten proposed the Barnes-Hut-SNE (BH-SNE)~\cite{VanDerMaaten:2014:BH-SNE}, which reduces the complexity of the algorithm to $O(N\log(N))$ for both the similarity computations and the objective function minimization.
More specifically, in the BH-SNE approach the similarity computations are seen as a $k$-nearest neighborhood graph computation problem, which is obtained using a Vantage-Point Tree~\cite{Yianilos:1993:VPTree}.
The minimization of the objective function is then seen as an $N$-body simulation, which is solved by applying the Barnes-Hut algorithm~\cite{Barnes:1986:BarnesHutAlgo}.

In our previous work~\cite{Pezzotti:2015:AtSNE}, we observed that the computation of the $k$-nearest neighborhood graph for high-dimensional spaces using the Vantage-Point Tree is affected by the curse of dimensionality, limiting the efficiency of the computation. 
To overcome this limitation, we proposed the Approximated-tSNE (A-tSNE) algorithm~\cite{Pezzotti:2015:AtSNE}, where approximated $k$-nearest neighborhood graphs are computed using a forest of randomized KD-trees~\cite{Muja:2014:Flann}.
Moreover, A-tSNE adopts the novel Progressive Visual Analytics paradigm~\cite{Stolper:2014:ProgressiveVisualAnalytics,Fekete:2016::PVA}, allowing the user to observe the evolution of the embedding during the minimization of the objective function.
This solution enables a user-driven early termination of the algorithm.
t-SNE-CUDA~\cite{chan2018t} is a CUDA implementation of the Approximated-tSNE algorithm. For computing the high-dimensional neighborhood, it uses the GPU library FAISS~\cite{JDH17}. A tree structure based on the BH-SNE is implemented in CUDA to compute the repulsive forces.
While the technique allows for a fast computation of the embedding, the application is limited to NVIDIA hardware, greatly limiting its application. Furthermore, like BH-SNE, the resulting embedding remains an approximation of the t-SNE embedding.

A similar observation on the benefit of using approximated computations was later made by Tang et al. that led to the development of the LargeVis technique~\cite{Tang:2016:LargeVis}.
LargeVis uses random projection trees~\cite{Dasgupta:2008:RandomPrjTrees} followed by a kNN-descent procedure~\cite{Dong:2011:kNNDescent} for the computation of the similarities and a different objective function that is minimized using a Stochastic Gradient Descent approach~\cite{Kiefer:1952:SGD}.
Despite the improvements, both the A-tSNE and LargeVis tools still suffer from long computation times during the optimization that hinder interaction for large data sets.
Better performance is achieved by the UMAP algorithm~\cite{McInnes:2018:UMAP}, which provides a different formulation of the dimensionality-reduction problem as a cross-entropy minimization between topological representations.
Computationally, UMAP follows LargeVis very closely and adopts a kNN-descent procedure~\cite{Dong:2011:kNNDescent} and stochastic gradient-descent minimization of the objective function.

A different approach is taken in the Hierarchical Stochastic Neighbor Embedding algorithm (HSNE)~\cite{Pezzotti:2016:HSNE}.
HSNE efficiently builds a hierarchical representation of the manifolds and embeds only a subset of the initial data that represents an overview of the available manifolds.
The user can ``drill-in'' the hierarchy by requesting more detailed embeddings that reveal smaller clusters of data points.
While HSNE allows scalability of the analysis to large data sets by the generation and user-guided exploration of multiple embeddings, it does not address the acceleration of the computation of single embeddings.

The techniques presented so far do not take advantage of the dimensionality of the target domain.
As a matter of fact, t-SNE is mostly used for data visualization in two-dimensional scatterplots, while the techniques introduced in this section so far are general and can be used in target domains of any dimensionality.
Based on this observation, Kim et al. introduced the PixelSNE technique~\cite{Kim:2016:PixelSNE}, where the points are not embedded in a continuous 2D space, but rather in a discretized space corresponding to the pixels used to display the scatterplot.
The optimization is performed using an $N$-body simulation approach, which is similar to the one employed by BH-SNE.
In order to compute embeddings that faithfully preserve high-dimensional neighborhoods, a large number of pixels must be used, often much larger than the display's resolution.
In addition, it hampers the scalability of the technique, requiring many hours to compute embeddings containing more than a million points.

In our work, we take advantage of the two-dimensional domain in which the embedding resides and we propose an efficient way to minimize the t-SNE objective function.
Contrary to PixelSNE, we only discretize the two-dimensional space for the computation of the repulsive forces presented in Equation~\ref{eq:C_gradient_n_body_detail}. 
We developed a linear-complexity approach implemented using GPGPU as a desktop and client-side browser application.
This is an improvement over t-SNE-CUDA, which can only be run on NVIDIA GPUs\revDeleted{, while providing higher quality embeddings.}\revNew{. Even though their computation of the embedding is faster, our technique produces embeddings that match more closely to the high-dimensional space.}
Compared to BH-SNE and PixelSNE, our technique computes embeddings with more than a million points in just a few minutes instead of several hours, while providing better preservation of high-dimensional similarities.

\section{Linear complexity t-SNE minimization}
\label{sec:main}

In this section, we present our approach to minimizing the t-SNE objective function as presented in Equation~\ref{eq:KL}. The main idea consists in rewriting the gradient presented in Equation~\ref{eq:C_gradient_n_body} such that it relies on a scalar field $\mathcal{S}$ and a vector field $\mathcal{V}$ in the 2D embedding domain.
These fields can be computed in linear time on the GPU and queried in constant time. Therefore, the complexity of the algorithm is reduced from quadratic to linear.

\begin{figure*}[!t]
\centering
\includegraphics[width=1\linewidth]{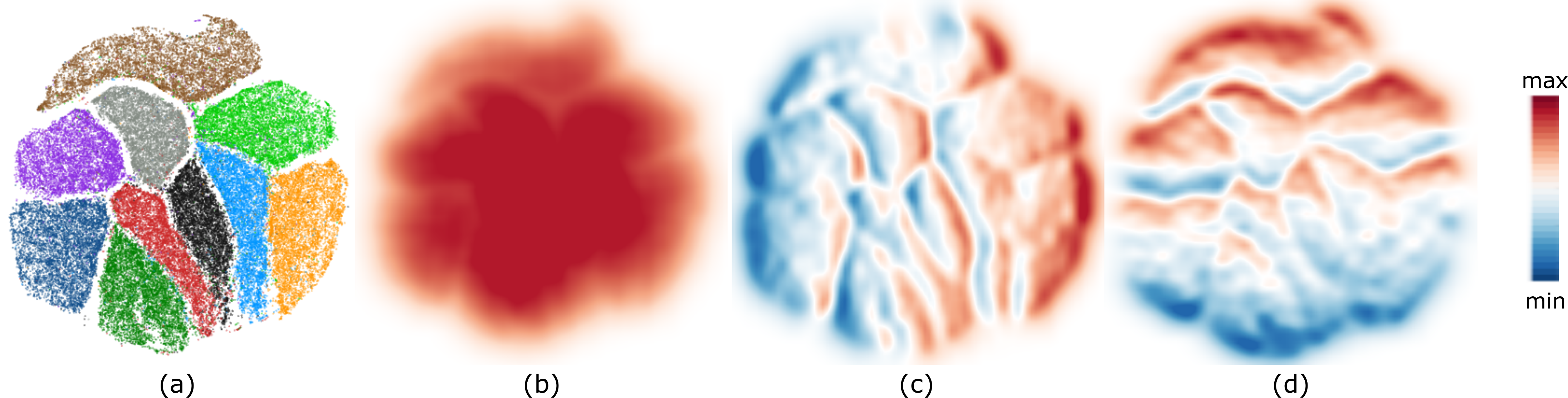}
\caption{\textbf{Fields} used in our approach. (a) The MNIST dataset contains images of handwritten digits and is embedded in a 2-dimensional space.
The minimization of the objective function is computed in linear time by making use of a scalar field $\mathcal{S}$ (b) and a 2-dimensional vector field $\mathcal{V}$, where (c-d) show the horizontal and vertical components respectively.
The fields are computed on the GPU by drawing properly designed mathematical kernels using the additive blending function of the rendering pipeline.
The rest of the minimization is treated as a series of tensor computations that are computed on the GPU.
}
\label{fig:teaser}
\end{figure*}

\subsection{Field-based computation of the gradient}
The gradient of the objective function has the same form as in regular t-SNE:
\begin{equation} \label{eq:our}
\frac{\delta C}{\delta \mathbf{y}_i} = 4 ( \hat{F}^{\text{attr}}_i-\hat{F}^{\text{rep}}_i),
\end{equation}
with attractive and repulsive forces acting on every point $\mathbf{x}_i \in X$.
We denote the forces with a $\wedge$ to distinguish them from their original counterparts.
We rewrite the equation of the gradient in the form of a scalar field $\mathcal{S}$ and a vector field $\mathcal{V}$:

\begin{equation} 
\label{eq:field_S}
\mathcal{S}(\mathbf{p}) = \sum_{i}^N \left(1+||\mathbf{y}_i - \mathbf{p}||^2 \right)^{-1}, \mathcal{S}: \mathbb{R}^2\Rightarrow \mathbb{R}
\end{equation}
\begin{equation}
\label{eq:field_V}
\mathcal{V}(\mathbf{p}) = \sum_{i}^N \left(1+||\mathbf{y}_i - \mathbf{p}||^2 \right)^{-2} (\mathbf{y}_i - \mathbf{p}), \mathcal{V}: \mathbb{R}^2\Rightarrow\mathbb{R}^2
\end{equation}

Intuitively, $\mathcal{S}$ represents the density of the points in the embedding space, according to the \revDeleted{t-Student distribution}\revNew{Student's t-distribution}, and it is used to compute the normalization of the joint probability distribution $Q$.
An example of the field $\mathcal{S}$ is shown in Figure~\ref{fig:teaser}b.
The vector field $\mathcal{V}$ represents the directional repulsive force applied to the entire embedding space.
An example of $\mathcal{V}$ is presented in Figures~\ref{fig:teaser}c and d, where the horizontal and vertical gradient components are visualized separately.
If a point in the embedding resides in the red area of Figure~\ref{fig:teaser}c, it will be pushed a certain amount to the right in the current iteration of the gradient descent, while a point in the blue area will be pushed to the left. Similarly for the vertical component, see Figure~\ref{fig:teaser}d, a point will be pushed either up, for red areas, or down for blue ones.
We describe the construction of $\mathcal{S}$ and $\mathcal{V}$ in Section~\ref{sec:fields_computation}. For now, we assume these fields are given, and we present how the gradient of the objective function is derived from $\mathcal{S}$ and $\mathcal{V}$.

For the attractive forces, we adopt the restricted neighborhood contribution as presented in the Barnes-Hut-SNE technique~\cite{VanDerMaaten:2014:BH-SNE}.
The rationale of this approach is that, by imposing a fixed perplexity on the Gaussian kernel, only a limited number of neighbors effectively apply an attractive force on any given point (see Equations~\ref{eq:P_j|i} and~\ref{eq:perplexity}).
Therefore we limit the number of contributing points to some multiple of the chosen perplexity. This approach reduces the computational and memory complexity of the computation of the attractive forces to $O(N)$, since the size of the neighborhood $k$ is several orders of magnitude lower than $N$, $k \ll N$.

\begin{equation} \label{eq:our_attractive}
\hat{F}^{\text{attr}}_i = \hat{Z}\sum_{l \in \text{kNN}(i) }  p_{il}q_{il}(\mathbf{y}_i-\mathbf{y}_l)
\end{equation}

The computation of the normalization factor $Z$, as it is presented in Equation~\ref{eq:Z_normalization_factor}, has computational complexity $O\left(N^2\right)$.
In our approach, we compute $\hat{Z}$ by consulting the scalar field $\mathcal{S}$ in constant time, giving us a complexity of $O\left(N\right)$.

\begin{equation} \label{eq:S_field}
\hat{Z} = \sum_{l=1}^N\left(\mathcal{S}(\mathbf{y}_l)-1\right)
\end{equation}

Note that the formulation of $Z$ and $\hat{Z}$ is identical but, since $\mathcal{S}$ is computed in linear time, computing $\hat{Z}$ also has linear complexity.
$\hat{Z}$ does not depend on the point $\mathbf{y}_i$ for which we are computing the gradient. Therefore, $\hat{Z}$  needs to be computed just once, cached, and then used at each iteration of the gradient descent for all points.

The repulsive force assumes the following form
\begin{equation} \label{eq:our_repulsive}
\hat{F}^{\text{rep}}_i = \mathcal{V}(\mathbf{y}_i) / \hat{Z},
\end{equation}
where the value of the vector field $\mathcal{V}$ in the location identified by the coordinates $\mathbf{y}_i$ is normalized by $\hat{Z}$.
Similar to $\hat{Z}$, $\hat{F}^{\text{rep}}$ has an equivalent formulation as $F^{\text{rep}}$ but with computational and memory complexity equal to $O(N)$.
So far, we assumed that $\mathcal{S}$ and $\mathcal{V}$ are computed in linear time and queried in constant time.
In the next section, we present how the rasterization pipeline is used to compute an approximation of the $\mathcal{S}$ and $\mathcal{V}$ fields. In Section~\ref{sec:impl}, two ways to implement the proposed approach are given.

\begin{figure}[!t]
\centering
\includegraphics[width=1\linewidth]{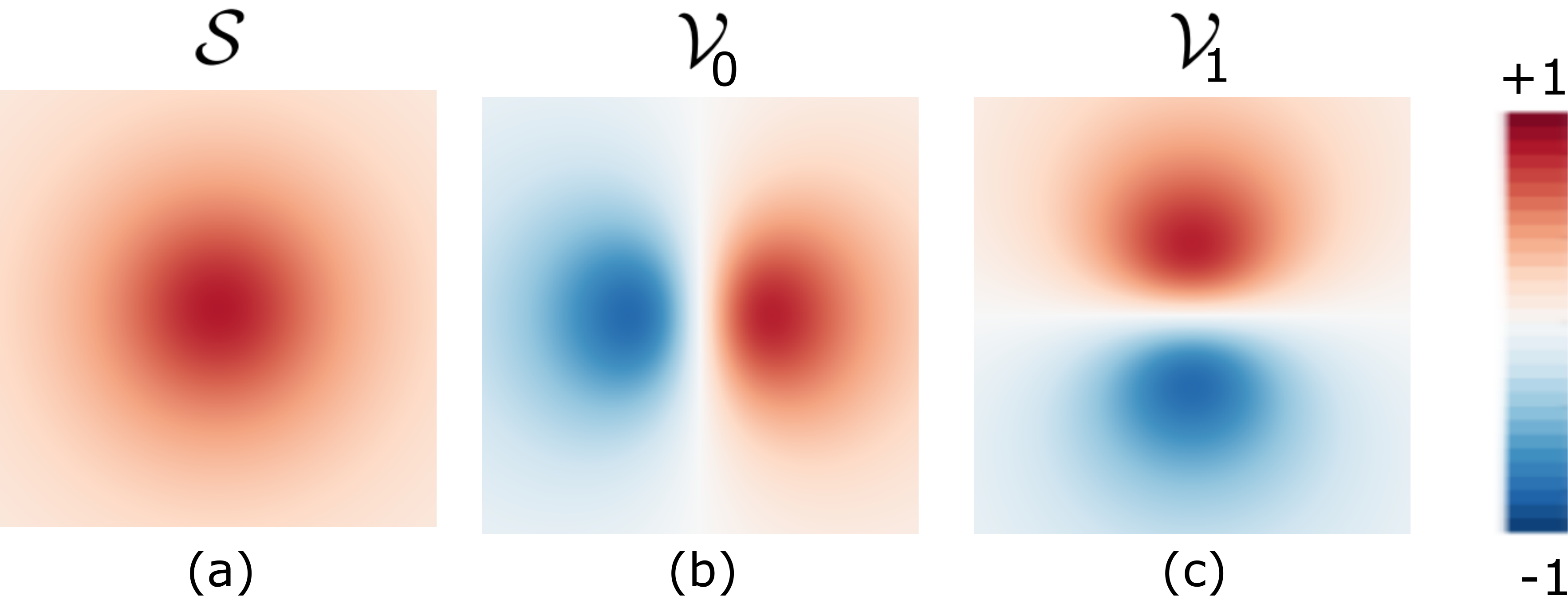}
\caption{\textbf{Functions} drawn over each embedding point to approximate the scalar field $\mathcal{S}$ and the 2-dimensional vector field $\mathcal{V}$.
}
\label{fig:kernels}
\end{figure}

\subsection{Computation of supporting fields}
\label{sec:fields_computation}

Our approach to the computation of the fields resembles an approach used for Kernel Density Estimation~\cite{Rosenblatt:1956:KDE}, which has applications in visualization~\cite{Lampe:2011:interactiveKDE} and non-parametric clustering~\cite{Hollt:2016:Cytosplore}.
In this setting, given a number of points, the goal is to estimate a two-dimensional probability density function. This is achieved by superimposing a Gaussian kernel, whose $\sigma$ has to be estimated, over every data point. Summing the contributions of all points in a given location or pixel in the embedding gives us the probability density function in a given location.


In KDE methods, the 2D kernel density is estimated efficiently on the GPU because of the quasi-limited support of the kernels, i.e., having values almost equal to zero if they are sufficiently far away from the origin. A good approximation of the density function is then achieved by drawing a quad at the location of each sample, which contains a precomputed texture or evaluates the kernel for each covered pixel~\cite{Lampe:2011:interactiveKDE,Bezerra:2008:dynamicGrouping}. By using additive blending, i.e., by summing the values in every pixel, the resulting output approximates the desired density function.

In our context, we want to compute $\mathcal{S}$ and $\mathcal{V}$ as shown in equations~\ref{eq:field_S} and~\ref{eq:field_V}. These equations can also be seen as a summation of kernels $S$ and $V$ as defined in the following equations:

\begin{equation}
\label{eq:field_S_summand}
\mathcal{S}(\mathbf{p}) = \sum_{i}^N S(\mathbf{y}_i-\mathbf{p}),\quad S(\mathbf{d}) = \left(1+||\mathbf{d}||^2 \right)^{-1}
\end{equation}
\begin{equation}
\label{eq:field_V_summand}
\mathcal{V}(\mathbf{p}) = \sum_{i}^N V(\mathbf{y}_i-\mathbf{p}),\quad V(\mathbf{d}) = \left(1+||\mathbf{d}||^2 \right)^{-2} ( \mathbf{d})
\end{equation}

The presented kernels $S$ and $V$ are stored in a texture and are presented in Figure~\ref{fig:kernels}.
The kernels have a limited function support, making it indeed very similar to the Kernel Density Estimation case discussed before.
As the fields $\mathcal{S}$ and $\mathcal{V}$ are a summation of the aforementioned kernels, we can compute an approximation of the fields by additively rendering these per-point kernel textures at the locations of each of the points in the embedding.
\begin{figure*}[!t]
\centering
\includegraphics[width=0.9\linewidth]{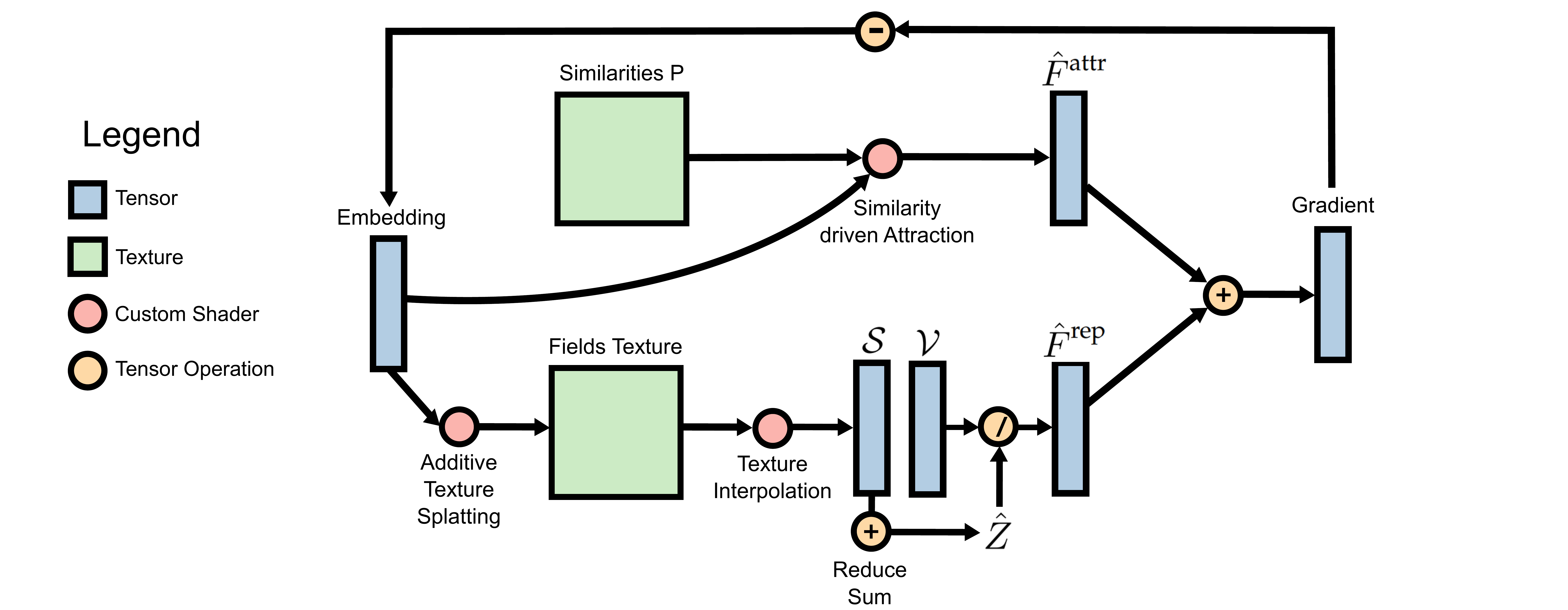}
\vspace{-3mm}
\caption{\textbf{Computational workflow} of our approach. On the lower side of the chart, the computation of the repulsive forces is presented. The fields texture is generated by the additive texture splatting presented in Section~\ref{sec:repulsive_forces}. The values of $\mathcal{S}$ and $\mathcal{V}$ are obtained through texture interpolation and are used to compute the repulsive forces. The attractive forces are computed in a custom shader that takes as input the similarities $P$ and the embedding. The gradient of the objective function is then computed using both forces and is used to update the embedding.
\vspace{-3mm}
}
\label{fig:workflow}
\end{figure*}

The resulting 3-channel texture, an example of which is presented in Figures~\ref{fig:teaser}b-d, represents the scalar field $\mathcal{S}$ and the vector field $\mathcal{V}$. Fetching the value of $\mathcal{S}$ and $\mathcal{V}$ for a point $\mathbf{y}_i$ then corresponds to extracting the interpolated value at the point's position in the field textures.

Contrary to the Kernel Density Estimation case, where the size of the quads changes according to the $\sigma$ chosen for the Gaussian kernel, our functions must have a fixed support in the embedding space. This is dictated by the fact that we are optimizing Equation~\ref{eq:KL}, a change of the quad size corresponds to a change in the low-dimensional distribution characterizing the points.
Therefore, the resolution of the texture influences the quality of the approximation but not the overall shape of the fields.
To achieve linear complexity, we define the resolution of the aggregate field texture according to the size of the embedding.
The number of pixels that are covered by the textures presented in Figure~\ref{fig:kernels} is kept constant.
This is achieved by changing the size of the target texture in the embedding space. A ratio $\rho$ between the diameter of the embedding and the texture resolution is fixed.
Hence, every data point updates the value of a constant number of pixels in the target texture equal to $\rho^2$.
This solution leads to $O(N \rho^2)$ complexity for the computation of the fields, and we empirically found $\rho = 0.5$ to be a good compromise between the fidelity of the resulting fields and the computation time required.
Since $ \rho^2 \ll N$, the resulting computational complexity is $O(N)$.
Note that, by being adaptive to the texture size, no parameter tuning is required.
A potential limitation is the maximum embedding size as defined by the OpenGL standard. In practice, this does not pose a limit since the embeddings size does.

\section{Implementations}
\label{sec:impl}

In this section we explain how the ideas presented in the previous section are implemented both for the browser as part of TensorFlow.js and for the desktop as part of the open source High-Dimensional Inspector (HDI) library~\cite{Pezzotti:2017:HDI}. Two different approaches are presented: one that makes use of the rasterization pipeline, and one that uses compute shaders. 

\subsection{Rasterization Approach}
In this section, we present an implementation that heavily makes use of the rasterization pipeline of modern GPUs. Rasterization is the task of converting a series of geometric primitives, most commonly triangles, into a series of pixels that form a raster image.
Contrary to the common application of rasterization in computer graphics, i.e., rendering of geometric scenes, here we associate each pixel with an atomic computation used for minimizing the t-SNE loss function.
These are the computation of the attractive forces given the similarity distribution $P$ (Section~\ref{sec:attractive_forces}), the computation of the fields used for computing the repulsive forces (Section~\ref{sec:repulsive_forces}) and subsequently the updating of the embedding (Section~\ref{sec:pnt_update}). 

\subsubsection{Attractive Forces}
\label{sec:attractive_forces}
Computation of the attractive forces, shown in the upper portion of Figure~\ref{fig:workflow}, is performed by measuring the sum of the contribution of every neighboring point in the high-dimensional space. The neighborhoods are encoded in the joint probability distribution $P$ which is stored in a sparse matrix. $P$ can be computed ahead of time, for example using an approximated $k$-nearest-neighborhood algorithm~\cite{Muja:2014:Flann,Dong:2011:kNNDescent,Dasgupta:2008:RandomPrjTrees} or by the HSNE technique~\cite{Pezzotti:2016:HSNE}. We use existing techniques here, and do not provide any contribution.


\subsubsection{Repulsive Forces}
\label{sec:repulsive_forces}
We achieve linear complexity for the computation of the repulsive forces by making use of the rasterization pipeline innate in graphics cards. For the browser implementation we make use of the WebGL API and for the desktop implementation we use standard OpenGL.

In order to form the field textures we start with a randomly initialized t-SNE embedding. Centered on each of the points in the embedding, a quad is rendered. We apply a texture to the quad whose R color channel contains $S(\mathbf{p})$ from Equation ~\ref{eq:field_S_summand} and whose G and B color channels contain $V(\mathbf{p})$ in each dimension from Equation~\ref{eq:field_V_summand}. By enabling additive blending these splatted textures will add up to an approximation of the  $\mathcal{S}$ and $\mathcal{V}$ fields. The approximated fields are stored in another floating-point RGB texture whose resolution is proportional to the size of the embedding space. The ratio between the two is defined by the parameter $\rho$ introduced in Section~\ref{sec:fields_computation}. 
The degree of approximation is controlled by the resolution of the aggregate field texture and the resolution of the kernel texture. 

To query the field values for a specific point in the embedding, we sample the field value at the point's position using bilinear texture interpolation. This operation is natively supported in the GPU and very efficient. The normalization factor $\hat{Z}$ is obtained by summing all the elements in the tensor with the interpolated values of $\mathcal{S}$.
This summation is performed as a reduction operation on the graphics card.
Note that $\hat{Z}$ is computed once and cached, hence Equation~\ref{eq:our_repulsive} is computed by simply dividing the interpolated field value by the cached $\hat{Z}$.

\begin{figure*}[!t]
\centering
\includegraphics[width=0.90\linewidth]{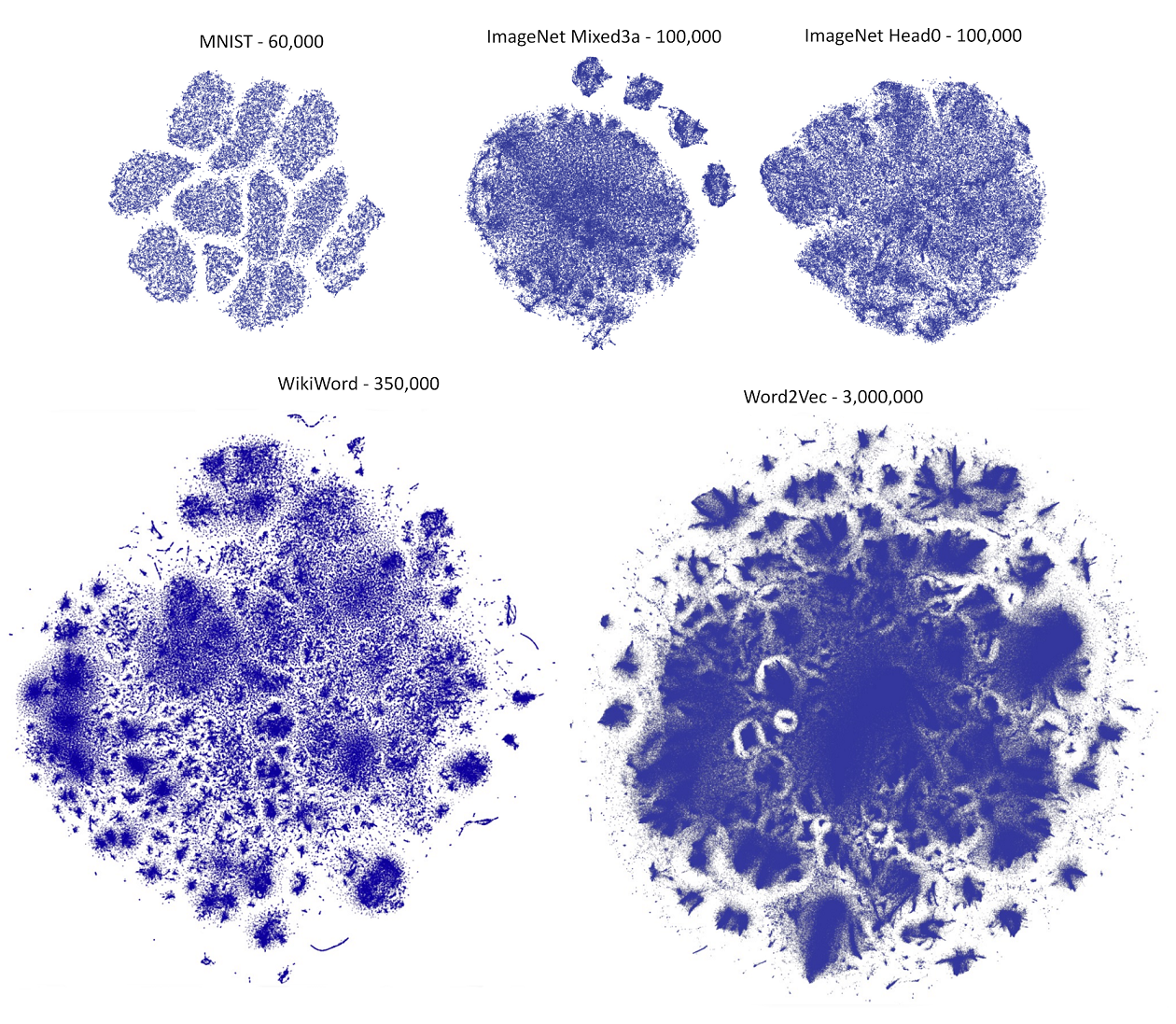}
\caption{\textbf{Embeddings} of the MNIST, ImageNet Mixed3a, ImageNet Head0, WikiWord and Word2Vec datasets generated by our technique.}
\label{fig:embeddings}
\end{figure*}

\subsubsection{Updating the points}
\label{sec:pnt_update}
The remaining computational steps are computed as tensor, i.e., matrix, operations as defined in toolkits like TensorFlow.js.
$\hat{F}^\text{rep}$ is obtained by dividing the interpolated values of $\mathcal{V}$ by $\hat{Z}$, and the gradient of the objective function is obtained by adding the attractive forces $\hat{F}^\text{attr}$.
The gradient is then applied to the embedding modifying the position of the points according to the gradient. 
Figure~\ref{fig:workflow} shows an overview of our approach.
Green squares represent textures containing the computed fields or the similarity matrix $P$, while blue rectangles represent tensors.
Operations are represented by circles. 
More specifically, red circles are custom operations that are implemented specifically for our technique.
Orange circles are tensor operations that are commonly available in TensorFlow.js or in the HDI library.

\subsection{Compute Shader Approach}
Implementations of our approach are available for both the web and desktop. These implementations are broadly applicable due to their limited feature requirements. However, as the computation of the algorithm is essentially reduced to a series of tensor operations, it lends itself very well to execution using one of the GPGPU APIs available. \revDeleted{This allows skipping the rasterization pipeline completely. This has the advantage that, when a lot of kernel splats overlap each other, i.e., when there is a high degree of overdraw, additively blending all the splats one by one can be quite costly. In particular, when the function support is large and substantial overdraw occurs, the algorithm will slow down significantly.}
\revNew{In the rasterization approach, many splats might overlap with each other. In particular, when the function support of the t-distribution is increased for more accurate embeddings, this simultaneously results in more overlapping splats. With additive blending enabled, this results in a high degree of overdraw, which can be quite costly.}
For this reason we have developed another implementation of the previously described algorithm.
Instead of splatting textures to obtain the fields, here, we calculate the fields in a compute shader in the following manner.

For each pixel in the output field we calculate the influence of per-point kernels on this pixel. If the point lies further away from the current pixel in embedding space than the given function support, the point is ignored. The complexity of this operation is  $O\left( N\:Px \right)$ where $Px$ represents the number of pixels used for the output field. In practice, our solution behaves very linearly, since the maximum number of pixels affected is much lower than the number of points in reasonably sized data sets. This means, that the function support can be unbounded with negligible loss of performance\revNew{, thereby resulting in even more accurate embeddings}. This can also be done in the rasterization approach, however, it would result in extreme overdraw and have a significant impact on performance.

\section{Evaluation}
\label{sec:res}


In order to assess the efficacy of the proposed technique we evaluate the computational costs and quality of the embedding using three metrics. First, we record the execution time of the minimization process over 1000 iterations.
Secondly, we evaluate the quality of the resulting embedding by using the reached Kullback--Leibler divergence. Kullback--Leibler is the objective function of the t-SNE algorithm. This metric shows how well the objective function is optimized by the different techniques.
We also compute the Nearest-Neighbor Preservation (NNP) metric as described by Venna et al. \cite{Venna:2010:InformationRetrievalDR} and implemented by Ingram and Munzner~\cite{ingram2015dimensionality}.
It measures how well small neighborhoods in the high-dimensional space are preserved during the dimensionality reduction.
The main benefit of such a metric is its independence from the objective function optimized by the t-SNE algorithm. In order to measure the NNP accurately it is important that the gradient descent has fully converged. We chose 1000 iterations for the MNIST and ImageNet datasets and 5000 iterations for the WikiWord and Word2Vec datasets to guarantee full convergence for the different data sizes.

We compare the results of our technique (i.e., GPGPU-SNE) with the results obtained from the Barnes-Hut-SNE~\cite{Barnes:1986:BarnesHutAlgo} and the t-SNE algorithm without computational improvements~\cite{VanDerMaaten:2008:tSNE}.
Both implementations are written in C++, support multi-threaded computations and are openly available in the High-Dimensional-Inspector (HDI) library~\cite{Pezzotti:2017:HDI}.
For Barnes-Hut-SNE, we provide results for two different values of its $\theta$ parameter. This parameter controls the trade-off between speed and accuracy of the algorithm. A value of $\theta=0.5$ sacrifices accuracy slightly for the benefit of a significant performance boost, and is often chosen as the default value. A value of $\theta=0.1$ prioritizes generating embeddings closer to those produced by original t-SNE, but at considerable execution time cost.
Moreover, we provide a comparison with the t-SNE-CUDA algorithm~\cite{chan2018t} for a value of $\theta=0.0$ and $0.5$.

We expect that our implementation outperforms BH-SNE in time as well as quality of the embeddings. 
Our approach is fundamentally a different method of acceleration compared to t-SNE-CUDA. \revDeleted{Our method does not require a NVIDIA GPU and, therefore, can be used to create embeddings in the web-browser.}\revNew{Our method does not rely on the CUDA API and can therefore be used to create embedding in a web-browser.} Concerning performance, we expect t-SNE-CUDA to be similar or better concerning the computational costs, but lower in quality since it is an acceleration based on the approximation of BH-SNE.

\begin{figure*}[!t]
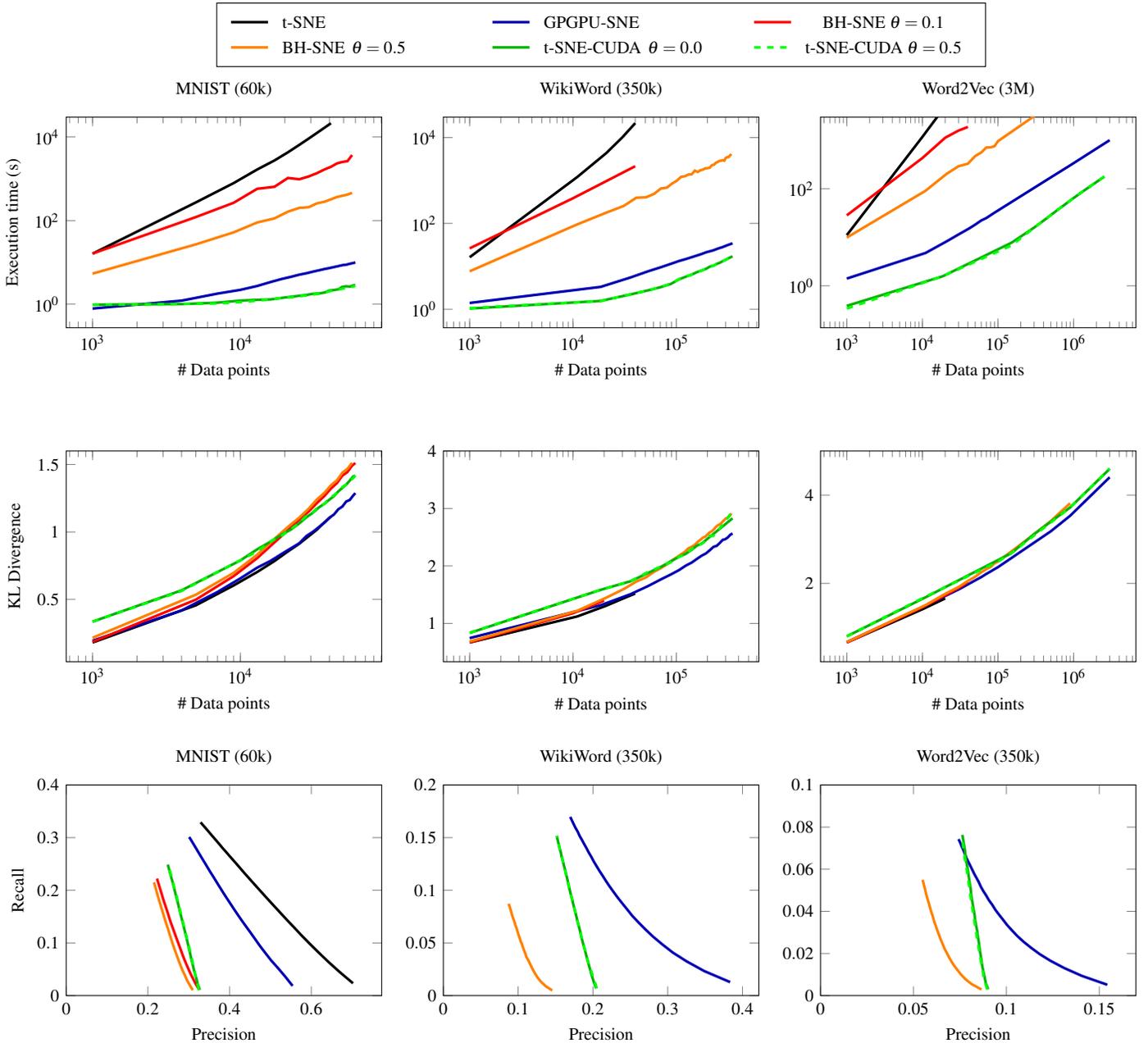

\begin{tikzpicture}

    \begin{groupplot}[group style={group size= 3 by 3,horizontal sep= 1.0cm,vertical sep=2.0cm},width=0.37\textwidth, height=5.0cm]
	    \nextgroupplot[title=MNIST (60k),xmode = log,ymode = log,ylabel={Execution time (s)},xlabel={\# Data points},scaled x ticks = false, x tick label style={/pgf/number format/fixed,/pgf/number format/1000 sep = \thinspace},every axis/.append style={font=\footnotesize},legend style={font=\tiny},ymax = 30000,]
			\input{charts/ET_MNIST.tex}
			
                \coordinate (top) at (rel axis cs:0,1);

        \nextgroupplot[title=WikiWord (350k),xmode = log,ymode = log,xlabel={\# Data points},scaled x ticks = false, x tick label style={/pgf/number format/fixed,/pgf/number format/1000 sep = \thinspace},every axis/.append style={font=\footnotesize},legend style={font=\tiny},ymax = 30000]
			\input{charts/ET_WikiWord.tex}
			
        \nextgroupplot[title=Word2Vec (3M),xmode = log,ymode = log,xlabel={\# Data points},scaled x ticks = false, x tick label style={/pgf/number format/fixed,/pgf/number format/1000 sep = \thinspace},every axis/.append style={font=\footnotesize},legend style={font=\tiny},ymax = 3000]
			\input{charts/ET_Word2Vec.tex}
			
			\coordinate (top) at (rel axis cs:0,1);

			\nextgroupplot[ylabel={KL Divergence},xmode = log,xlabel={\# Data points},scaled x ticks = false, x tick label style={/pgf/number format/fixed,/pgf/number format/1000 sep = \thinspace},every axis/.append style={font=\footnotesize},legend style={font=\tiny},ymax = 1.6,]
			\input{charts/KL_MNIST.tex}
			
                \coordinate (top) at (rel axis cs:0,1);

        \nextgroupplot[xlabel={\# Data points},xmode = log,scaled x ticks = false, x tick label style={/pgf/number format/fixed,/pgf/number format/1000 sep = \thinspace},every axis/.append style={font=\footnotesize},legend style={font=\tiny},ymax = 4.0]
			\input{charts/KL_WikiWord.tex}
			
        \nextgroupplot[xlabel={\# Data points},xmode = log,scaled x ticks = false, x tick label style={/pgf/number format/fixed,/pgf/number format/1000 sep = \thinspace},every axis/.append style={font=\footnotesize},legend style={font=\tiny},ymax = 5.0]
			\input{charts/KL_Word2Vec.tex}
                \coordinate (bot) at (rel axis cs:1,0);


		\nextgroupplot[title=MNIST (60k), ylabel={Recall},xlabel={Precision},scaled x ticks = false, x tick label style={/pgf/number format/fixed,/pgf/number format/1000 sep = \thinspace},every axis/.append style={font=\footnotesize},legend style={font=\tiny},xmin=0,ymin=0,ymax = 0.4,]
			\input{charts/PR_MNIST.tex}
                \coordinate (top) at (rel axis cs:0,1);
                
        \nextgroupplot[title=WikiWord (350k), xlabel={Precision},scaled x ticks = false, x tick label style={/pgf/number format/fixed,/pgf/number format/1000 sep = \thinspace}, y tick label style={/pgf/number format/fixed},every axis/.append style={font=\footnotesize},legend style={font=\tiny},xmin=0,ymin=0,ymax = 0.2]
			\input{charts/PR_WikiWord.tex}
			
        \nextgroupplot[title=Word2Vec (350k), xlabel={Precision},scaled x ticks = false, x tick label style={/pgf/number format/fixed,/pgf/number format/1000 sep = \thinspace}, y tick label style={/pgf/number format/fixed},every axis/.append style={font=\footnotesize},legend style={font=\tiny},xmin=0,ymin=0,ymax = 0.1]
			\input{charts/PR_Word2Vec.tex}
                \coordinate (bot) at (rel axis cs:1,0);
                \coordinate (bot) at (rel axis cs:1,0);

    \end{groupplot}
\path (top|-current bounding box.north)--
      coordinate(legendpos)
      (bot|-current bounding box.north);
\matrix[
    matrix of nodes,
    anchor=south,
    draw,
		column 2/.style={nodes={text width=3cm,align=left}},
		column 4/.style={nodes={text width=3cm,align=left}}
  ]at([yshift=1ex]legendpos)
  {
	  \ref{plots:plot1}& {\footnotesize t-SNE} &[5pt]
	  \ref{plots:plot2}& {\footnotesize GPGPU-SNE} &[5pt]
	  \ref{plots:plot3}& {\footnotesize BH-SNE $\theta=0.1$} &[5pt]\\
      \ref{plots:plot4}& {\footnotesize BH-SNE $\theta=0.5$} &[5pt]
      \ref{plots:plot5}& {\footnotesize t-SNE-CUDA $\theta=0.0$}&[5pt]
      \ref{plots:plot6}& {\footnotesize t-SNE-CUDA $\theta=0.5$} &[5pt]\\
  };
\end{tikzpicture}

    \caption{\textbf{Results} of the experiments on the MNIST, WikiWord and Word2Vec datasets for the t-SNE, Barnes-Hut-SNE, t-SNE-CUDA and our approach. The first row shows the evolution of the execution time with increasingly bigger subsets of the dataset. The second row shows how well the objective function is fulfilled, while the third row shows the Nearest-Neighborhood Preservation (NPP). \revDeleted{Our technique is up to two orders of magnitude faster than BH-SNE and provides better embedding quality.}\revNew{Our technique is up to two orders of magnitude faster than Barnes-Hut-SNE and provides higher quality embeddings compared to Barnes-Hut-based techniques.}}
    \label{fig:error_approximation}

\end{figure*}

\subsection{Datasets}
We have chosen five commonly used datasets to illustrate the applicability of our technique to both small and large amounts of high-dimensional data.
First, we use the \textbf{MNIST} dataset. 
It consists of 60k labeled grayscale images of handwritten digits (compare Figure~\ref{fig:teaser}a). 
Each image is represented as a 784 dimensional vector, corresponding to the gray values of the pixels in the image.
The MNIST data is often used to validate non-linear dimensionality reduction techniques. 
As a matter of fact, it clearly contains 10 different manifolds, one for each digit.
Moreover, the manifolds are non-linear, hence linear dimensionality-reduction techniques such as PCA are not able to reconstruct the manifolds.

\begin{table}[h]
\caption{\textbf{Datasets} used for the evaluation.\vspace{-2mm}} \label{tab:datasettable}
\begin{tabular}{ |c|c|c| } 
 \hline
 Dataset & Number of points & Number of dimensions \\
 \hline\hline
 MNIST-60000 & 60000 & 768 \\ 
 \hline
 WikiWord & 350000 & 300 \\ 
 \hline
 GoogleNews & 3000000 & 300 \\ 
 \hline
 ImageNet Mixed3a & 100000 & 256 \\ 
 \hline
 ImageNet Head0 & 100000 & 128 \\ 
 \hline
\end{tabular}
\end{table}

The \textbf{WikiWord} and \textbf{GoogleNews} datasets contain words, which are associated with a vector representation.
These vector representations are algorithmically generated by processing large text corpora, often through a deep neural network~\cite{LeCun:2015:DeepLearning} and by requiring that words that occur in similar contexts share a similar representation.
The shapes associated with each word present interesting characteristics for latent semantic analysis~\cite{landauer1998introduction}.
As an example, it is shown that simple summation and subtraction of the vectors representing the words $King - Man + Woman$, as produced by the GloVe model~\cite{pennington2014glove}, is very similar to the vector representation associated with the word $Queen$.
Non-linear dimensionality reduction is often used in systems for the analysis of such word representations~\cite{liu2018visual,hohman2018visual,chen2018evaluating}.

Finally, we present two different datasets obtained by collecting the activations of different layers in a deep neural network (DNN)~\cite{LeCun:2015:DeepLearning} on the validation set of the ImageNet dataset~\cite{Alex:2012:ImageNet}\footnote{The datasets can be created for an arbritary activation layer using the following Colab Notebook: \url{https://colab.research.google.com/github/tensorflow/lucid/blob/master/notebooks/activation-atlas/activation-atlas-simple.ipynb}}.
The resulting embeddings shed a light on the internal computations performed by the deep neural network, the Google Inception~\cite{szegedy2015going} in this case. 
Images, or image patches, that are close in the embedding are considered similar by the DNN~\cite{Pezzotti:2018:DeepEyes}.
Recently, an increasing number of web-based tools, like the Activation Atlas ~\cite{carter2019activation} or Tensorboard, have been proposed to better understand and improve DNNs through dimensionality reduction techniques such as t-SNE or UMAP.

\subsection{Results}
In Figure~\ref{fig:error_approximation}, we show the results of the experiments for the chosen datasets. All experiments are conducted on an Intel Core i7-4820K Processor, with 4 physical cores (8 threads) @ 3.70 Ghz.
The machine has 16GB of DDR3 RAM, and an NVIDIA GeForce GTX Titan GPU with 2688 CUDA cores @ 837 Mhz and 6GB of GDDR5 memory. All experiments run fit in the main memory available and have no interaction with disk during the optimization process.

To better highlight the behaviour of the algorithms with increasing dataset sizes, we run the algorithm on a random subset of the data with a growing number of data points for each of the experiments.
The first row of charts in Figure~\ref{fig:error_approximation} shows the execution time of the various algorithms plotted against the number of data points in the subsampled dataset. Note that a logarithmic scale is used for both the vertical and horizontal axes.

Our technique significantly cuts back on execution time compared to Barnes-Hut-SNE and t-SNE. For the MNIST dataset, t-SNE takes two days to complete the iterations. BH-SNE with  $\theta=0.1$ takes one hour and with  $\theta=0.5$ takes around 8 minutes. While our technique computes the embedding in just 16 seconds. This is a reduction on the cost of the gradient descent in the range of orders of magnitude. For the other datasets it becomes infeasible to run the first two algorithms as they would take many days to execute. It is possible to run BH-SNE $\theta=0.5$ on the WikiWord dataset, but the computation takes more than an hour, while our technique computes the embedding in a mere 35 seconds.
t-SNE-CUDA outperforms our technique by a factor \revDeleted{of $x5$}\revNew{in the range of $x2$ to $x5$}. This can be explained by the highly-optimized code enabled by the CUDA implementation.

The second row examines the KL-divergence of the final embeddings from their original high-dimensional counterparts. And the last row shows the Nearest Neighborhood Preservation of all the embeddings, presented as a precision/recall plot.

In comparison to other optimization methods our technique produces a better, i.e., lower KL-divergence at data sets of non-trivial size. A likely explanation for this is that as the datasets get larger, the domain of the embedding expands but this expansion is not linear in the number of points. Therefore, the embedding will get progressively more dense, which is unfavourable for the Barnes-Hut approximation, which is also used by the t-SNE-CUDA. Approximations of the forces applied by distant points will become coarser as more of them are lumped together. Consequently this lowers the accuracy of the algorithm. This results in embeddings where the objective function cannot be effectively minimized, hence resulting in lower nearest-neighbor preservation.
This observation is confirmed by the results presented in the third row.
A similar observation can be made for the t-SNE-CUDA algorithm. Here, even higher KL-divergence can be observed for lower numbers of data points in the embedding. Speed is traded in favour of quality in producing the final embedding.

\begin{figure}[!t]
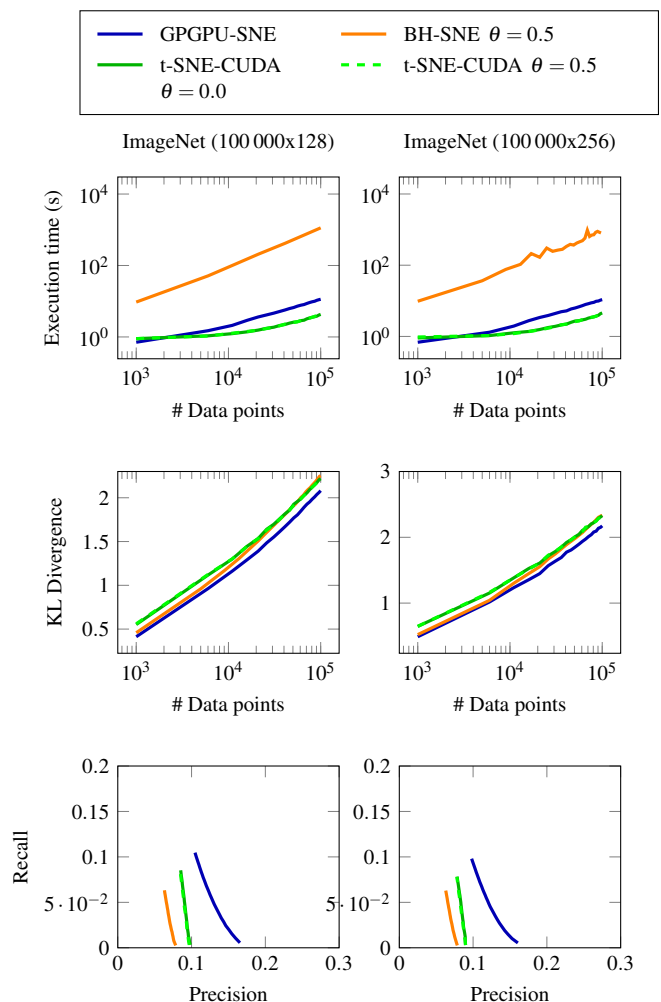

\begin{tikzpicture}

    \begin{groupplot}[group style={group size= 2 by 3,horizontal sep= 0.8cm,vertical sep= 1.5cm},width=0.25\textwidth, height=4.0cm]
	    \nextgroupplot[title=ImageNet (100\,000x128),xmode = log,ymode = log,ylabel={Execution time (s)},xlabel={\# Data points},scaled x ticks = false, x tick label style={/pgf/number format/fixed,/pgf/number format/1000 sep = \thinspace},every axis/.append style={font=\footnotesize},legend style={font=\tiny},ymax = 30000,]
			\input{charts/ET_ImageNet128.tex}
			
                \coordinate (top) at (rel axis cs:0,1);

        \nextgroupplot[title=ImageNet (100\,000x256),xmode = log,ymode = log,xlabel={\# Data points},scaled x ticks = false, x tick label style={/pgf/number format/fixed,/pgf/number format/1000 sep = \thinspace},every axis/.append style={font=\footnotesize},legend style={font=\tiny},ymax = 30000]
			\input{charts/ET_ImageNet256.tex}
			
			\coordinate (top) at (rel axis cs:0,1);

			\nextgroupplot[ylabel={KL Divergence},xmode = log,xlabel={\# Data points},scaled x ticks = false, x tick label style={/pgf/number format/fixed,/pgf/number format/1000 sep = \thinspace},every axis/.append style={font=\footnotesize},legend style={font=\tiny},ymax = 2.3,]
			\input{charts/KL_ImageNet128.tex}
			
                \coordinate (top) at (rel axis cs:0,1);

        \nextgroupplot[xlabel={\# Data points},xmode = log,scaled x ticks = false, x tick label style={/pgf/number format/fixed,/pgf/number format/1000 sep = \thinspace},every axis/.append style={font=\footnotesize},legend style={font=\tiny},ymax = 3.0]
			\input{charts/KL_ImageNet256.tex}

                \coordinate (bot) at (rel axis cs:1,0);


		\nextgroupplot[ylabel={Recall},xlabel={Precision},scaled x ticks = false, x tick label style={/pgf/number format/fixed,/pgf/number format/1000 sep = \thinspace},every axis/.append style={font=\footnotesize},legend style={font=\tiny},xmin=0,ymin=0,ymax = 0.2,xmax= 0.3]
			\input{charts/PR_ImageNet128.tex}
                \coordinate (top) at (rel axis cs:0,1);
			
        \nextgroupplot[xlabel={Precision},scaled x ticks = false, x tick label style={/pgf/number format/fixed,/pgf/number format/1000 sep = \thinspace},every axis/.append style={font=\footnotesize},legend style={font=\tiny},xmin=0,ymin=0,ymax = 0.2,xmax= 0.3]
			\input{charts/PR_ImageNet256.tex}
                \coordinate (bot) at (rel axis cs:1,0);
                \coordinate (bot) at (rel axis cs:1,0);

    \end{groupplot}
\path (top|-current bounding box.north)--
      coordinate(legendpos)
      (bot|-current bounding box.north);
\matrix[
    matrix of nodes,
    anchor=south,
    draw,
		column 2/.style={nodes={text width=2cm,align=left}},
		column 4/.style={nodes={text width=3cm,align=left}}
  ]at([yshift=1ex]legendpos)
  {
	  \ref{plots:plot2}& {\footnotesize GPGPU-SNE} &[5pt]
      \ref{plots:plot4}& {\footnotesize BH-SNE $\theta=0.5$}&[5pt]\\
      \ref{plots:plot5}& {\footnotesize t-SNE-CUDA $\theta=0.0$}&[5pt]
      \ref{plots:plot6}& {\footnotesize t-SNE-CUDA $\theta=0.5$}\\
  };
\end{tikzpicture}

    \caption{\textbf{Results} of the experiments on the ImageNet datasets for Barnes-Hut-SNE, t-SNE-CUDA and our approach.}
    \label{fig:error_approximation1}
\end{figure}

In the last row of Figure~\ref{fig:error_approximation}, we present the nearest-neighbor preservation for the different data sets.
For each point, we examine a neighborhood of $k$ points in the high and low-dimensional space. For every value from $k=1$ to $k=30$ we compute the true positive $T$, defined as the points that belong to both neighborhoods.
From this, we compute precision as $T/k$, while recall is defined as $T/30$.
The values of precision and recall for each value of $k$ form a precision/recall curve for every point.
The precision/recall curve for the entire embedding is obtained by averaging the curves of every point in the dataset.
\revDeleted{In the case of the MNIST dataset, the metric is computed on embeddings of the full dataset generated by each of the algorithms. For the other datasets it is computed on embeddings of a subset of 60,000 points.} Since t-SNE and BH-SNE with $\theta=0.1$ take days to compute on these datasets, it becomes infeasible to calculate the metric for all datasets. We provide it for the MNIST dataset to give an indication of the relationship between the techniques. In addition, for the 3-million data point Word2Vec dataset calculating the metric would take more than a week. Therefore, we compute it on a 350k subset of the dataset, which also allows the curve for Barnes-Hut-SNE to be presented.
We see that our technique has a significant advantage over the Barnes-Hut-SNE and t-SNE-CUDA algorithm, as it presents a high Precision/Recall curve in all measured datasets.
Figure~\ref{fig:error_approximation1} shows the results on the ImageNet datasets for our technique, BH-SNE with $theta=0.5$ and t-SNE-CUDA with $theta=0.0$ and $0.5$. The results confirm the previous analysis, showing that our technique beats the BH-SNE by almost two orders of magnitude. 
t-SNE-CUDA is faster by a factor of approximately $x3$ on the full dataset, requiring less than 4 seconds while our approach computes the embeddings in 11 seconds.
Our solution, however, shows lower KL-divergence and better precision and recall than both BH-SNE and t-SNE-CUDA.


\section{Conclusion}
In this work, we presented a novel approach for the optimization of the objective function of t-SNE that scales to large datasets.
We provided a reformulation of the gradient equations of the objective function that includes a scalar and a vector field. These fields represent the point density and the directional repulsive forces in the embedding space. Our approach relies on modern graphics hardware to efficiently compute these fields, obtaining linear complexity in the number of points compared to the quadratic complexity of the non-accelarated t-SNE.

In our experiments, we observe that our implementation outperforms the Barnes-Hut-SNE algorithm by several orders of magnitude.
Besides the faster optimization, our technique is better at minimizing the objective function than all other acceleration methods, i.e., having a lower Kullback-Leibler divergence, and provides better Nearest-Neighbor Preservation. 
t-SNE-CUDA outperforms our method in computational times, but produces lower quality embeddings, and relies on NVIDIA GPUs, which limits its applicability.

We provide two implementations of our technique.
The first one is available in the High-Dimensional Inspector library.
The library, which can be found at the following link \url{https://github.com/Nicola17/High-Dimensional-Inspector}, is a C++ library used by several visual-analytics applications such as Cytosplore~\cite{Hollt:2016:Cytosplore,hollt2017cyteguide,vanUnen:2017:HSNECytof}.
The second implementation is released as part of TensorFlow.js and can be found on GitHub at the following address: \url{https://github.com/tensorflow/tfjs-tsne}.

As future work, we want to explore how our implementation can be integrated in Progressive Visual Analytics systems~\cite{Fekete:2016::PVA,turkay2018progressive}, such as tools for the analysis of Deep Neural Networks. For example, the Embedding Projector~\footnote{\url{https://projector.tensorflow.org}}, TensorBoard~\footnote{\url{https://www.tensorflow.org/programmers\_guide/summaries\_and\_tensorboard}} and DeepEyes~\cite{Pezzotti:2018:DeepEyes}.
A limitation of the presented technique is that a graphics card is required in order to run the algorithm, which potentially restricts its applicability. In addition, our technique shares the intrinsic problems of t-SNE, such as a limited ability to reveal global relationships in the data. Therefore, we are interested in extending our approach to other techniques that better address this problem, such as UMAP~\cite{McInnes:2018:UMAP} and HSNE~\cite{Pezzotti:2016:HSNE}.
To conclude, we believe that our technique is an enabler for more interactive high-dimensional data analysis, in particular thanks to the possibility of optimizing embeddings directly in the browser.


\acknowledgments{
The authors wish to thank the Google AI team PAIR for supporting the development of the TensorFlow.js implementation.
This  work  received  funding  through  the STW Project 12720, VAnPIRe.}

\bibliographystyle{abbrv-doi}

\bibliography{template}
\end{document}